\documentclass[runningheads]{llncs}

\usepackage{eccv}

\usepackage{eccvabbrv}

\usepackage{graphicx}
\usepackage{booktabs}
\usepackage{colortbl}
\usepackage{xcolor}
\usepackage{booktabs}
\usepackage[table]{xcolor}
\usepackage[accsupp]{axessibility}  

\usepackage[backref=page]{hyperref}
\usepackage{orcidlink}

\begin{document}

\title{VOS-Agent: The 1st Place Solution for the 8th LSVOS Challenge (MOSEv2 Track)} 

\titlerunning{The 1st Solution for 8th LSVOS Challenge MOSEv2 Track}

\author{Canyang Wu\inst{1} \and Jinrong Zhang \inst{1} \and Xusheng He\inst{1} \and Ce Bian\inst{1} \and Xianjing Han\inst{2} \and Jianlong Wu\inst{1,3}}

\authorrunning{C. Wu \etal}

\institute{Harbin Institute of Technology, Shenzhen \and Nanyang Technological University  \and Shenzhen Loop Area Institute 
	\\
	Team: HITsz-Dragon
}


\maketitle

\begin{abstract}
  Complex video object segmentation requires robust target propagation
  under severe occlusion, disappearance and reappearance. Although SAM3 provides strong
  promptable mask propagation, a uniform inference path remains unreliable
  for tiny targets with insufficient visual evidence and semantic-dominated
  targets whose identities depend on explicit attributes. To this end, we present
  VOS-Agent, a collaborative framework that retains SAM3 as
  the shared dense segmentation module and conditionally activates
  specialized agents according to target characteristics. A Target
  Perception and Routing Agent assigns each sequence to a regular, tiny,
  or semantic-dominated route. Tiny targets are supported by a Visual Tracking Agent through confidence-aware box prompts,
  while semantic-dominated targets are handled by an MLLM-based Semantic
  Agent through description-guided localization and candidate verification. On the MOSEv2 test set, VOS-Agent achieves 69.82\% on the official
  $\mathcal{J}\&\dot{\mathcal{F}}$ metric and ranks first in the MOSEv2
  Track of the 8th LSVOS Challenge at ECCV 2026.
  
\end{abstract}

\section{Introduction}
\label{sec:intro}
\begin{figure}[tb]
	\centering
	\includegraphics[height=4cm]{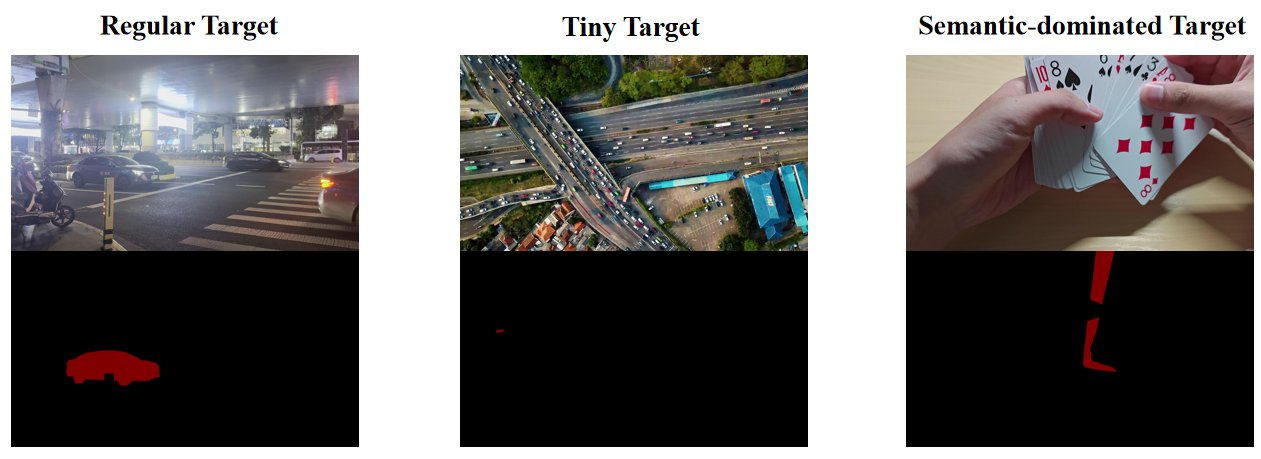}
	\caption{Visualization of the three target types in the MOSEv2 dataset, highlighting the challenges posed by tiny targets and
		semantic-dominated targets in complex environments. Tiny targets provide insufficient visual evidence for stable long-range correspondence while semantic-dominated targets remain difficult to indentity as mutiple nearby instances share similiar low-level appearance.
	}
	\label{fig:3_target_types}
\end{figure}

The 8th Large-scale Video Object Segmentation (LSVOS) Challenge aims to advance video object
segmentation toward increasingly complex and realistic scenarios.
The challenge includes three tracks: Complex Video Object Segmentation
(MOSEv2), Text-based Referring Motion Expression Video Segmentation
(MeViSv2-Text), and Audio-based Referring Motion Expression Video
Segmentation (MeViSv2-Audio)~\cite{lsvos2026}.
The MOSEv2 track focuses on tracking and segmenting objects under
challenging real-world conditions~\cite{ding2025mosev2}, while the
MeViSv2-Text and MeViSv2-Audio tracks investigate motion-expression-guided
video segmentation using textual and spoken descriptions,
respectively~\cite{ding2025mevis}.
This paper focuses on the MOSEv2 track.

Semi-supervised video object segmentation (VOS) aims to delineate one or more target objects throughout a video, given their segmentation masks in an initial frame. Recent researches have substantially improved mask propagation and long-term association on established VOS benchmarks~\cite{cheng2022xmem,cheng2024putting,ravi2025sam,ding2025sam2long,videnovic2025distractor,carion2026sam}. However, high performance on videos containing salient and relatively isolated objects does not necessarily translate to robustness in unconstrained environments. MOSE and its more challenging successor, MOSEv2, were introduced to evaluate VOS under crowded scenes, severe occlusion, object disappearance and reappearance, small or camouflaged targets, adverse illumination and weather conditions, and visually similar distractors~\cite{ding2023mose,ding2025mosev2}. 

A dominant line of VOS research improves temporal correspondence and memory construction. XMem organizes complementary memory stores for efficient long-term propagation~\cite{cheng2022xmem}, while Cutie introduces object-level memory reading to reduce matching noise in distractor-rich scenes~\cite{cheng2024putting}. SAM2 provides a general framework for prompt-conditioned mask propagation~\cite{ravi2025sam}, and subsequent methods improve its robustness through multi-path memory search or distractor-aware memory management~\cite{ding2025sam2long,videnovic2025distractor}. SAM3 further integrates concept-conditioned detection with video mask propagation, providing a strong foundation for promptable video segmentation~\cite{carion2026sam}.

Despite these advances, complex VOS still presents heterogeneous failure modes that cannot always be addressed by a single uniform inference path. In MOSEv2, as shown in \cref{fig:3_target_types}, tiny objects may occupy only a few pixels and provide insufficient visual evidence for stable long-range correspondence. They are therefore particularly vulnerable to background clutter, occlusion, and accumulated localization errors. Other targets have sufficient spatial extent but remain difficult to identify because multiple nearby instances share similar low-level appearance. Their identities may instead depend on higher-level attributes such as text, logos, symbols, colors, or distinctive clothing. We refer to targets whose identity is primarily determined by such explicit semantic attributes as \emph{semantic-dominated targets}.

Despite the strong promptable segmentation and temporal propagation
capabilities of foundation models such as SAM3, these two failure modes
remain difficult to resolve through a uniform inference path. For tiny
targets, insufficient visual evidence limits reliable localization and
long-term propagation. For semantic-dominated targets, low-level visual
correspondence alone is often inadequate for preserving identity among
similar instances. At their core, both failures arise from the same
limitation: SAM3 propagates the target primarily through prompt-conditioned
visual features and temporal memory. When the target representation is
either severely under-resolved or visually ambiguous, the tracker lacks
sufficient evidence to determine where the target is or which instance
should be preserved.

To this end, we propose VOS-Agent, a task-oriented collaborative multi-agent framework for complex video object segmentation. VOS-Agent comprises four components: a Target Perception and Routing Agent, a SAM3 Segmentation Agent, a Visual Tracking Agent, and an MLLM-based Semantic Agent. The Perception and Routing Agent analyzes the initial target scale and semantic distinctiveness and assigns each sequence to a regular, tiny, or semantic-dominated processing route. Regular targets are handled directly by the SAM3 Agent. For tiny targets, the Visual Tracking Agent uses the initial target crop as a visual template and collaborates with SAM3 through confidence-aware bounding-box prompts~\cite{chen2025sutrack}. For semantic-dominated targets, the Semantic Agent constructs a discriminative description of the target, performs language-guided localization, and compares candidate regions before returning the selected prompt to SAM3~\cite{zhang2026advancing}. The specialized agents operate primarily at the target-localization and identity-reasoning level, whereas the SAM3 Agent converts their guidance into dense pixel-level masks.

Overall, VOS-Agent provides a modular and training-free solution for 
complex video object segmentation. By routing heterogeneous targets to 
specialized processing paths, the framework preserves the strong 
pixel-level mask generation capability of SAM3 while complementing it 
with visual-exemplar tracking for tiny targets and semantic identity 
reasoning for semantic-dominated targets.  VOS-Agent achieves a $\mathcal{J}\&\mathcal{F^{*}}$ score of 
\textbf{69.82\%} on the MOSEv2 test set and ranks \textbf{1st} in the MOSEv2 Track of the
8th LSVOS Challenge at ECCV 2026.

\section{Solution}
\label{sec:sol}
\subsection{Problem Formulation}
\label{sec:formulation}

Given a video sequence
$\mathcal{V}=\{I_t\}_{t=1}^{T}$ and the ground-truth mask
$M_1$ of a target object in the first frame, semi-supervised
video object segmentation aims to predict a sequence of masks
$\{\widehat{M}_t\}_{t=2}^{T}$ that consistently delineates the
same target throughout the video. 

\subsection{Framework Overview}
\label{sec:overview}

\begin{figure}[tb]
	\centering
	\includegraphics[height=6.5cm]{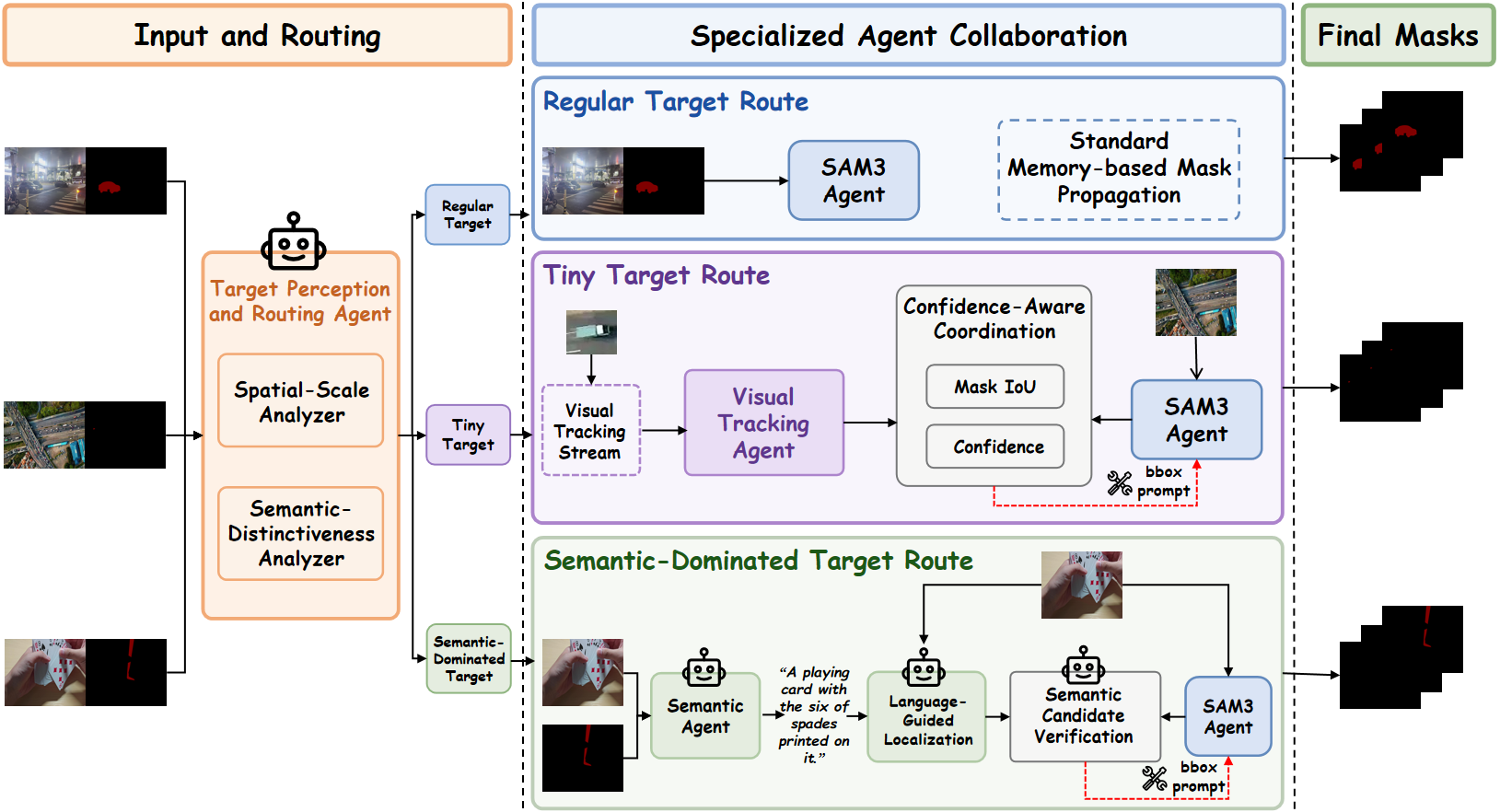}
	\caption{Overview of the proposed VOS-Agent framework. Given the first frame and its target mask, the Target Perception and Routing Agent categorizes the target as regular, tiny, or semantic-dominated according to its spatial scale and semantic distinctiveness. Regular targets are processed directly by the SAM3 Agent. For tiny targets, the Visual Tracking Agent recursively estimates the target location and provides a corrective box prompt when its high-confidence prediction disagrees with SAM3. For semantic-dominated targets, the Semantic Agent performs description-guided localization and candidate verification to preserve target identity. 
	}
	\label{fig:overview}
\end{figure}

As illustrated in \cref{fig:overview}, VOS-Agent consists of four core components: a Target
Perception and Routing Agent, a SAM3 Segmentation Agent, a Visual
Tracking Agent, and a Semantic Agent. Given the first
frame and its target mask, the Routing Agent assigns the target to
one of three processing routes. Regular targets are processed directly by the SAM3 Agent. For tiny targets, the Visual Tracking Agent recursively estimates the target location and provides a corrective box prompt when its high-confidence prediction disagrees with SAM3. For semantic-dominated targets, the Semantic Agent performs description-guided localization and candidate verification to preserve target identity. 

\subsection{Target Perception and Routing Agent}
\label{sec:routing_agent}

The Routing
Agent analyzes two complementary properties of the
reference target: its spatial scale and its semantic distinctiveness.

\subsubsection{Spatial-scale analysis.}
Let $A(M_1)$ denote the foreground area of the initial target mask,
and let $A(I_1)$ denote the image area. We define the normalized
target area as
\begin{equation}
	r_{\mathrm{area}}
	=
	\frac{A(M_1)}{A(I_1)}.
	\label{eq:area_ratio}
\end{equation}
A target is categorized as tiny when
$r_{\mathrm{area}} < \tau_{\mathrm{area}}$, where
$\tau_{\mathrm{area}}$ is selected on the validation set. This
criterion reflects the observation that very small targets provide
limited pixel-level evidence and are particularly susceptible to
feature degradation, background interference, and accumulated
localization errors.

\subsubsection{Semantic-distinctiveness analysis.}
For a non-tiny target, an MLLM examines the reference crop and its
surrounding context and determines whether explicit semantic
attributes are required to preserve target identity. The decision is
written as
\begin{equation}
	s_{\mathrm{sem}}
	=
	\mathcal{A}_{\mathrm{sem\mbox{-}cls}}(I_1,M_1),
	\qquad
	s_{\mathrm{sem}}\in\{0,1\}.
	\label{eq:semantic_decision}
\end{equation}
Examples of discriminative attributes include printed text, logos,
symbols, distinctive colors, playing-card identities, or unusual
clothing. A target is considered semantic-dominated when these
attributes are necessary to distinguish it from nearby instances
with similar low-level appearance.

The final routing decision is
\begin{equation}
	z =
	\begin{cases}
		\texttt{tiny},
		& r_{\mathrm{area}} < \tau_{\mathrm{area}}, \\[2mm]
		\texttt{semantic},
		& r_{\mathrm{area}} \geq \tau_{\mathrm{area}}
		\ \text{and}\ s_{\mathrm{sem}}=1, \\[2mm]
		\texttt{regular},
		& \text{otherwise}.
	\end{cases}
	\label{eq:routing_rule}
\end{equation}

\subsection{SAM3 Segmentation Agent}
\label{sec:sam_agent}

The SAM3 Agent serves as the shared dense segmentation and temporal
propagation module in all processing routes. 
\cite{carion2026sam}.

The initial mask $M_1$ initializes a target masklet and its
conditioning memory. For each subsequent frame $I_t$, the tracker
predicts the current target mask by combining the current-frame
features with a memory bank constructed from the initial conditioning
frame and confidently tracked previous frames. 

For regular targets, no additional prompt is introduced after
initialization, and SAM3 performs standard memory-based mask
propagation. For tiny and semantic-dominated targets, a specialized
agent may provide a bounding-box prompt on a selected frame. The
prompt is used to refine the current masklet and provide an updated
conditioning reference for subsequent propagation.

\subsection{Tracking Agent--SAM3 Agent Collaboration}
\label{sec:tracking_collaboration}

Tiny targets often provide insufficient pixel-level evidence for
stable long-term mask propagation. 
We therefore instantiate the Visual Tracking Agent with SUTrack,
which performs recursive single-object tracking using visual
templates and a local search region~\cite{chen2025sutrack}.

The tracker is initialized with the target box derived from the
first-frame mask:
\begin{equation}
	B_1^{\mathrm{trk}} = \mathcal{B}(M_1),
	\qquad
	Z^{\mathrm{sta}}
	=
	\operatorname{Crop}(I_1,B_1^{\mathrm{trk}}),
	\label{eq:tracking_initialization}
\end{equation}
where $Z^{\mathrm{sta}}$ denotes the static target template retained
throughout the sequence. SUTrack additionally maintains a dynamic
template $Z_{t-1}^{\mathrm{dyn}}$, which is updated online according
to its confidence-based template update strategy.

For each subsequent frame, a search region is cropped around the
previous tracking result:
\begin{equation}
	X_t
	=
	\operatorname{Crop}
	\left(
	I_t,
	\operatorname{Expand}
	\bigl(B_{t-1}^{\mathrm{trk}}\bigr)
	\right).
	\label{eq:tracking_search_region}
\end{equation}
The Tracking Agent then estimates the current target box and its
confidence from the static template, the dynamic template, and the
current search region:
\begin{equation}
	\bigl(B_t^{\mathrm{trk}},c_t^{\mathrm{trk}}\bigr)
	=
	\mathcal{A}_{\mathrm{trk}}
	\left(
	Z^{\mathrm{sta}},
	Z_{t-1}^{\mathrm{dyn}},
	X_t
	\right).
	\label{eq:tracking_prediction}
\end{equation}

In parallel, the SAM3 Agent propagates its target masklet and
produces a native mask $M_t^{\mathrm{sam}}$. Let
$B_t^{\mathrm{sam}}$ denote the tight bounding box enclosing this
mask. The agreement between the two agents is measured as
\begin{equation}
	q_t^{\mathrm{trk}}
	=
	\operatorname{IoU}
	\bigl(
	B_t^{\mathrm{sam}},
	B_t^{\mathrm{trk}}
	\bigr).
	\label{eq:tracking_agreement}
\end{equation}

A low agreement does not by itself indicate that the tracking result
is more reliable. We therefore accept the auxiliary tracking prompt
only when the two predictions disagree and the tracking confidence
is sufficiently high:
\begin{equation}
	\eta_t^{\mathrm{trk}}
	=
	\mathbb{I}
	\left[
	q_t^{\mathrm{trk}} < \tau_{\mathrm{iou}}
	\ \land\
	c_t^{\mathrm{trk}} \geq \tau_{\mathrm{conf}}
	\right].
	\label{eq:tracking_acceptance}
\end{equation}

When $\eta_t^{\mathrm{trk}}=1$, the tracking box is introduced as an
additional box prompt on frame $t$. SAM3 uses the accepted box prompt to refine the mask on the current
frame and to update its conditioning state for subsequent mask
propagation. If the tracking prompt is rejected, the native SAM3
prediction is retained.

The two agents consequently maintain complementary temporal states.
SUTrack recursively estimates target location from its previous
tracking result and template state, whereas SAM3 propagates a dense
mask through its video memory. The Tracking Agent intervenes only
when it provides a confident localization that is inconsistent with
the native SAM3 prediction. SAM3 then converts the accepted coarse
box into a dense mask and continues propagation from the corrected
conditioning state.

\subsection{Semantic Agent--SAM3 Agent Collaboration}
\label{sec:semantic_collaboration}

Semantic-dominated targets are generally large enough to be
segmented, but their identities may not be reliably preserved through
low-level visual correspondence alone. This situation commonly occurs
when several nearby instances share similar shapes, textures, or
colors, while the designated target is distinguished by explicit
attributes such as text, logos, symbols, or distinctive clothing.
We therefore employ an MLLM-based Semantic Agent to complement SAM3 with identity-level reasoning.

The Semantic Agent first generates a discriminative description from
the initial target:
\begin{equation}
	D
	=
	\mathcal{A}_{\mathrm{sem}}^{\mathrm{desc}}
	\bigl(I_1,M_1\bigr),
	\label{eq:target_description}
\end{equation}
where $D$ summarizes the visible attributes that distinguish the
designated target from same-category distractors. 

For each subsequent frame, the Semantic Agent performs
language-guided localization using the target description:
\begin{equation}
	B_t^{\mathrm{sem}}
	=
	\mathcal{A}_{\mathrm{sem}}^{\mathrm{loc}}
	\bigl(I_t,D\bigr),
	\label{eq:semantic_localization}
\end{equation}
where $B_t^{\mathrm{sem}}$ denotes the semantic localization result.
In parallel, the SAM3 Agent propagates its target masklet and produces
the native mask $M_t^{\mathrm{sam}}$. Let
$B_t^{\mathrm{sam}}$ denote the tight bounding box enclosing this
mask. Their spatial agreement is measured by
\begin{equation}
	q_t^{\mathrm{sem}}
	=
	\operatorname{IoU}
	\bigl(
	B_t^{\mathrm{sam}},
	B_t^{\mathrm{sem}}
	\bigr).
	\label{eq:semantic_agreement}
\end{equation}

When $q_t^{\mathrm{sem}}\geq\tau_{\mathrm{sem}}$, the two agents
provide consistent localization and the native SAM3 prediction is
retained. When their agreement falls below
$\tau_{\mathrm{sem}}$, the Semantic Agent performs explicit candidate
verification. We first extract the reference target and the two
current-frame candidates:
\begin{equation}
	\begin{split}
		R_1^{\mathrm{ref}}
		&=
		\operatorname{Crop}
		\bigl(I_1,\mathcal{B}(M_1)\bigr),\\
		R_t^{\mathrm{sam}}
		&=
		\operatorname{Crop}
		\bigl(I_t,B_t^{\mathrm{sam}}\bigr),\\
		R_t^{\mathrm{sem}}
		&=
		\operatorname{Crop}
		\bigl(I_t,B_t^{\mathrm{sem}}\bigr).
	\end{split}
	\label{eq:semantic_crops}
\end{equation}
The Semantic Agent then compares both candidates with the initial
target and its discriminative description:
\begin{equation}
	y_t^{\mathrm{sem}}
	=
	\mathcal{A}_{\mathrm{sem}}^{\mathrm{judge}}
	\left(
	R_1^{\mathrm{ref}},
	D,
	R_t^{\mathrm{sam}},
	R_t^{\mathrm{sem}}
	\right),
	\qquad
	y_t^{\mathrm{sem}}
	\in
	\{\texttt{sam},\texttt{semantic}\}.
	\label{eq:semantic_judgement}
\end{equation}
Here, $y_t^{\mathrm{sem}}=\texttt{semantic}$ indicates that the
language-guided candidate is judged to be more consistent with the
designated target than the native SAM3 candidate. The semantic
correction is accepted only when the two agents disagree and the
semantic candidate is selected:
\begin{equation}
	\eta_t^{\mathrm{sem}}
	=
	\mathbb{I}
	\left[
	q_t^{\mathrm{sem}} < \tau_{\mathrm{sem}}
	\ \land\
	y_t^{\mathrm{sem}}=\texttt{semantic}
	\right].
	\label{eq:semantic_acceptance}
\end{equation}

When $\eta_t^{\mathrm{sem}}=1$, the semantic bounding box is
introduced as an additional conditioning prompt on frame $t$ to refine the mask.

The Semantic Agent intervenes
only when the language-guided localization conflicts with the native
SAM3 prediction and is judged to better preserve the original target
identity. SAM3 then refines the accepted bounding-box prompt into a
dense mask and updates its conditioning state for subsequent
propagation.

\section{Experiment}
\label{sec:experiments}

\subsection{Dataset and Challenge Protocol}
\label{sec:dataset}

We evaluate VOS-Agent on the MOSEv2 track of the 8th Large-scale
Video Object Segmentation Challenge at ECCV
2026. MOSEv2 contains 5,024 videos, 10,074 annotated objects from
200 categories, and more than 701,000 high-quality masks
\cite{ding2025mosev2}. Compared with conventional VOS benchmarks,
it contains more frequent disappearance and reappearance events,
smaller and less conspicuous targets, severe occlusion, crowded
scenes, adverse weather, low-light environments, multi-shot
sequences, camouflaged objects, non-physical targets, and cases that
require external knowledge. The first-frame mask of each evaluated target is provided, while the
masks of the remaining frames must be inferred.

\subsection{Implementation Details}
\label{sec:implementation}

The SAM3 Agent is implemented using the official SAM3 model
\cite{carion2026sam}. The Target Perception and Routing Agent is instantiated with Qwen3.5-397B-A17B~\cite{team2026qwen3}. The Visual Tracking Agent is instantiated
with SUTrack~\cite{chen2025sutrack} and receives the RGB crop enclosed by the first-frame
mask as its visual exemplar. The Semantic
Agent is instantiated with Qwen3.5-397B-A17B~\cite{team2026qwen3}, which performs
target-description generation, language-guided localization, and
candidate comparison. 

\subsection{Main Results}
\begin{table}[t]
	\centering
	\caption{
		Quantitative comparison of VOS-Agent with existing solutions on the
		MOSEv2 test set. We report the overall
		$\mathcal{J}\&\dot{\mathcal{F}}$, region similarity $\mathcal{J}$,
		boundary accuracy $\dot{\mathcal{F}}$, and performance on
		disappearance ($\mathrm{d}$) and reappearance ($\mathrm{r}$) scenarios.
	}
	\label{tab:quantitative results}
	\setlength{\tabcolsep}{8pt}
	\renewcommand{\arraystretch}{1.15}
	\resizebox{\textwidth}{!}{
		\begin{tabular}{lccccccc}
			\hline
			\textbf{Participant}
			& $\mathbf{\mathcal{J}\&\dot{\mathcal{F}}}$
			& $\mathbf{\mathcal{J}}$
			& $\mathbf{\dot{\mathcal{F}}}$
			& $\mathbf{\mathcal{J}\&\dot{\mathcal{F}}_{\mathrm{d}}}$
			& $\mathbf{\mathcal{J}\&\dot{\mathcal{F}}_{\mathrm{r}}}$
			& $\mathbf{\mathcal{F}}$
			& $\mathbf{\mathcal{J}\&\mathcal{F}}$ \\
			\hline
			
			\rowcolor{gray!20}
			\textbf{HITsz-Dragon}
			& \textbf{69.82}
			& \textbf{68.21}
			& \textbf{71.43}
			& 79.12
			& \textbf{34.87}
			& \textbf{73.76}
			& \textbf{70.99}
			\\
			
			mmm
			&66.20
			&64.79
			&67.60
			&\textbf{81.57}
			&26.88
			&69.74
			&67.26
			\\
			
			kjeong
			&64.37
			&63.16
			&65.59
			&80.26
			&27.53
			&67.09
			&65.12
			\\
			
			\hline
		\end{tabular}
	}
\end{table}

\begin{table}[t]
	\centering
	\caption{
		Ablation study on the MOSEv2 test set. The best result in each
		column is shown in bold. The subscripts $\mathrm{d}$ and
		$\mathrm{r}$ denote disappearance and reappearance scenarios,
		respectively.
	}
	\label{tab:ablation}
	\setlength{\tabcolsep}{3.5pt}
	\renewcommand{\arraystretch}{1.15}
	\resizebox{\textwidth}{!}{
		\begin{tabular}{lccccccc}
			\toprule
			\textbf{Method}
			& $\mathbf{\mathcal{J}\&\dot{\mathcal{F}}}$
			& $\mathbf{\mathcal{J}}$
			& $\mathbf{\dot{\mathcal{F}}}$
			& $\mathbf{\mathcal{J}\&\dot{\mathcal{F}}_{\mathrm{d}}}$
			& $\mathbf{\mathcal{J}\&\dot{\mathcal{F}}_{\mathrm{r}}}$
			& $\mathbf{\mathcal{F}}$
			& $\mathbf{\mathcal{J}\&\mathcal{F}}$ \\
			\midrule
			
			SAM3 Agent
			& 62.49
			& 61.38
			& 63.59
			& 78.93
			& 25.31
			& 65.41
			& 63.40 \\
			
			Routing + Semantic Agent + SAM3 Agent
			& 67.07
			& 65.68
			& 68.47
			& \textbf{79.12}
			& 31.24
			& 70.35
			& 68.02 \\
			
			\rowcolor{gray!12}
			\textbf{Full VOS-Agent}
			& \textbf{69.82}
			& \textbf{68.21}
			& \textbf{71.43}
			& \textbf{79.12}
			& \textbf{34.87}
			& \textbf{73.76}
			& \textbf{70.99} \\
			
			\bottomrule
		\end{tabular}
	}
\end{table}

\begin{figure}[tb]
	\centering
	\includegraphics[height=10cm]{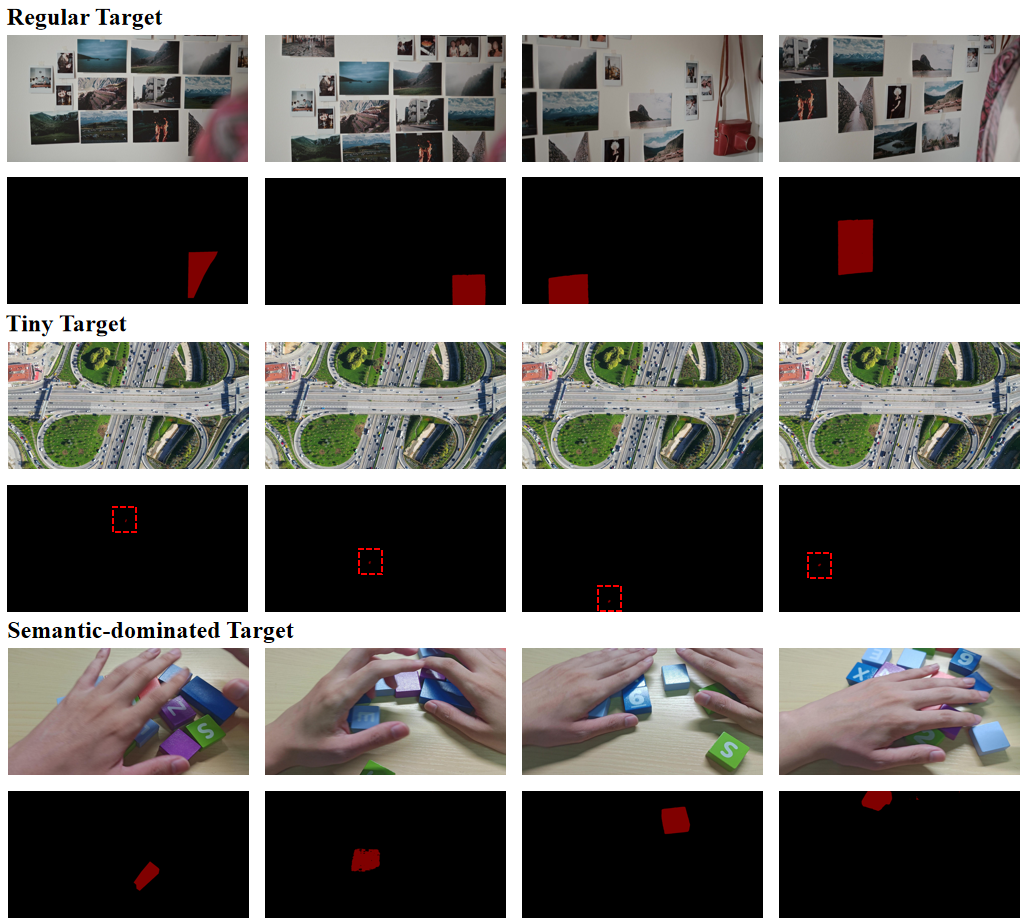}
	\caption{Qualitative visualization of segmentation results on the MOSEv2 test set. The visualizations demonstrate that the tracking and segmentation contours for both tiny and semantic-dominated targets remain clean, tight, and temporally stable, validating the efficacy of VOS-Agent.
	}
	\label{fig:qualitative_results}
\end{figure}

\cref{tab:quantitative results} reports the final performance of VOS-Agent on the MOSEv2 test set. Our method ranks 1st in the challenge leaderboard and achieves 69.82 on the official $\mathcal{J}\&\dot{\mathcal{F}}$ metric. This result suggests that routing heterogeneous targets to specialized tracking and semantic reasoning agents, while retaining SAM3 as the shared mask-propagation module, is effective for handling tiny targets, semantic-dominated targets, and frequent disappearance--reappearance patterns. 

As shown in \cref{fig:qualitative_results}, the qualitative results further illustrate the
behavior of VOS-Agent across different target types. For regular
targets, SAM3 produces accurate and temporally consistent masks without
additional intervention. For tiny targets, the Tracking Agent maintains
stable localization despite the extremely small object scale, allowing
SAM3 to recover compact masks across successive frames. For
semantic-dominated targets, the Semantic Agent preserves the identity of
the designated instance among visually similar objects and prevents
instance switching. These examples strongly demonstrate that the specialized agents provide
complementary guidance for challenging localization and identity
disambiguation cases.

\subsection{Ablation Study}
\label{sec:ablation}

To validate the effectiveness of each core agent, we evaluate three configurations directly on the official MOSEv2
test set: (1) use the SAM3 Agent alone and
applies the same propagation path to all targets; (2)
introduce the Target Perception and Routing Agent together with the
Semantic Agent; (3) the complete VOS-Agent further activates the Visual
Tracking Agent for tiny targets.

As shown in \cref{tab:ablation}, introducing target routing and
the Semantic Agent improves the primary
$\mathcal{J}\&\dot{\mathcal{F}}$ score from 62.49 to 67.07,
demonstrating the benefit of explicit identity reasoning for
semantic-dominated targets. Adding the Visual Tracking Agent further
raises the score to 69.82. Overall, VOS-Agent improves
$\mathcal{J}\&\dot{\mathcal{F}}$ by 7.33 points, indicating that the
tracking and semantic agents provide complementary localization cues
for heterogeneous target types.

\section{Conclusion}
In this report, we presented VOS-Agent, our solution to the MOSEv2
Track of the 8th LSVOS Challenge. VOS-Agent complements SAM3 with
category-specific tracking and semantic reasoning for tiny and
semantic-dominated targets, improving robustness to target loss,
identity ambiguity, and reappearance without fine-tuning the SAM3 backbone. Our method achieves 69.82\%
$\mathcal{J}\&\dot{\mathcal{F}}$ on the MOSEv2 test set and ranks
first in the challenge, showing that specialized agent collaboration
is an effective strategy for complex semi-supervised VOS.

%
%
\bibliographystyle{splncs04}
\bibliography{main}
\end{document}